\documentclass[12pt]{article}

\newcommand{\x}{\mathbf{x}}

\newcommand{\y}{\mathbf{y}}
\newcommand{\rs}{\mathbf{r}}
\newcommand{\bi}{\mathbf{I}}
\newcommand{\bz}{\mathbf{0}}
\usepackage{etoolbox}
\usepackage{bm}

\usepackage{arydshln}

\makeatletter
\def\mathcolor#1#{\@mathcolor{#1}}
\def\@mathcolor#1#2#3{%
	\protect\leavevmode
	\begingroup
	\color#1{#2}#3%
	\endgroup
}
\makeatother

\usepackage{url}
\usepackage{setspace}
\usepackage{algorithm}
\usepackage{algpseudocode}
\algrenewcommand\algorithmicrequire{\textbf{Input:}}
\algrenewcommand\algorithmicensure{\textbf{Output:}}

\usepackage[table,xcdraw]{xcolor}

\usepackage{graphicx}
\usepackage{multirow}

\graphicspath{ {./figures/} }

\usepackage{caption}
\usepackage{subcaption}

\usepackage{amsmath} 
\usepackage{amsfonts}
\usepackage{nicefrac}

\usepackage{hyperref}
\hypersetup{
	colorlinks=true,
	linkcolor=blue,
	filecolor=blue,      
	urlcolor=blue,
	citecolor=blue,
}

\usepackage[a4paper, total={6in, 8in}]{geometry}

\usepackage[symbol]{footmisc}

\usepackage{multirow}
\usepackage[numbers,sort&compress]{natbib}

\begin{document}
	\begin{center}	
\sf {\Large {\bfseries Res-MoCoDiff: Residual-guided diffusion models for motion artifact correction in brain MRI}} \\ 

Mojtaba Safari$ ^1 $, Shansong Wang$ ^1 $, Qiang Li$ ^1 $, Zach Eidex$ ^1 $, Richard L.J. Qiu$ ^1 $, Chih-Wei Chang$ ^1 $, Hui Mao$ ^2 $, and Xiaofeng Yang$ ^{1, \ddagger} $ \\
	
\end{center}
{$ ^1 $Department of Radiation Oncology and Winship Cancer Institute, Emory University, Atlanta, GA 30322, United States.\\
$ ^2 $Department of Radiology and Image Science and Winship Cancer Institute, Emory University, Atlanta, United States.\\
$ ^\ddagger  $Corresponding Author: email: \url{xiaofeng.yang@emory.edu}
}\\

\newpage

\begin{abstract}
	\noindent\textit{Objective.}Motion artifacts in brain MRI, mainly from rigid head motion, degrade image quality and hinder downstream applications. Conventional methods to mitigate these artifacts, including repeated acquisitions or motion tracking, impose workflow burdens. This study introduces Res-MoCoDiff, an efficient denoising diffusion probabilistic model specifically designed for MRI motion artifact correction.\textit{Approach.}Res-MoCoDiff exploits a novel residual error shifting mechanism during the forward diffusion process to incorporate information from motion-corrupted images. This mechanism allows the model to simulate the evolution of noise with a probability distribution closely matching that of the corrupted data, enabling a reverse diffusion process that requires only four steps. The model employs a U-net backbone, with attention layers replaced by Swin Transformer blocks, to enhance robustness across resolutions. Furthermore, the training process integrates a combined $\ell_1 + \ell_2$ loss function, which promotes image sharpness and reduces pixel-level errors. Res-MoCoDiff was evaluated on both an \textit{in-silico} dataset generated using a realistic motion simulation framework and an \noindent\textit{in-vivo} MR-ART dataset. Comparative analyses were conducted against established methods, including CycleGAN, Pix2pix, and a diffusion model with a vision transformer backbone (MT-DDPM), using quantitative metrics such as peak signal-to-noise ratio (PSNR), structural similarity index measure (SSIM), and normalized mean squared error (NMSE).\textit{Main results.} The proposed method demonstrated superior performance in removing motion artifacts across minor, moderate, and heavy distortion levels. Res-MoCoDiff consistently achieved the highest SSIM and the lowest NMSE values, with a PSNR of up to $41.91\pm2.94$ dB for minor distortions. Notably, the average sampling time was reduced to 0.37 seconds per batch of two image slices, compared with 101.74 seconds for conventional approaches.\textit{Significance.}Res-MoCoDiff offers a robust and efficient solution for correcting MRI motion artifacts, preserving fine structural details while significantly reducing computational overhead. Its speed and restoration fidelity underscore its potential for integration into clinical workflows, enhancing diagnostic accuracy and patient care.
\end{abstract}


\newpage 
\section{Introduction}

Magnetic resonance imaging (MRI) is a cornerstone of modern diagnostics, treatment planning, and patient follow-up, providing high-resolution images of soft tissues without the use of ionizing radiation. However, prolonged MRI acquisitions increase the likelihood of patient movement, leading to motion artifacts. These artifacts can alter the B$_0$ field, which results in susceptibility artifacts~\cite{https://doi.org/10.1002/acm2.14072}, and disrupt the \textit{k}-space readout lines, potentially violating the Nyquist criterion and causing ghosting and ringing artifacts~\cite{Zaitsev_2015}. As one of the most common artifacts encountered in MRI~\cite{Sreekumari217}, motion artifacts may compromise post-processing procedures such as image segmentation~\cite{Kemenczky2022} and target tracking in MR-guided radiation therapy~\cite{https://doi.org/10.1002/mp.16702}. Moreover, due to the limited availability of prospective motion correction techniques and the additional complexity they introduce, clinical workflows often resort to repeating the imaging acquisition. The severity and spatial distribution of these artifacts underscore a need for robust methods capable of effectively removing or significantly reducing motion artifacts without repeated imaging.

Traditional motion correction (MoCo) algorithms have primarily aimed at mitigating artifacts by optimizing image quality metrics such as entropy and image gradient~\cite{doi:10.1148/radiology.215.3.r00jn19904}, as well as by estimating the motion-corrupted \textit{k}-space lines~\cite{https://doi.org/10.1002/mrm.10093} and corresponding motion trajectories~\cite{https://doi.org/10.1002/mrm.24615}. Additional strategies include navigator echoes, external tracking devices, and sequence modifications that prospectively adjust acquisition to account for motion~\cite{Zaitsev_2015}. Retrospective approaches in the reconstruction domain, such as phase shift correction in \textit{k}-space, blind motion trajectory estimation, and compressed sensing with motion models, have also been widely explored~\cite{doi:10.1148/radiology.215.3.r00jn19904}. While these methods have shown clinical utility, their adoption is often limited by the need for raw \textit{k}-space data, which is not routinely stored in clinical archives, and by reconstruction pipelines that vary across scanner vendors, reducing reproducibility and generalizability~\cite{Tamir2025, https://doi.org/10.1002/mrm.29292}. 

In parallel, deep learning (DL) approaches have demonstrated superior performance in suppressing motion artifacts compared to conventional methods~\cite{10285512,DUFFY2021117756}. In particular, both supervised and unsupervised generative models based on generative adversarial networks (GANs) have been successfully employed for MRI motion artifact removal~\cite{https://doi.org/10.1002/mrm.27783, Safari_2024_pmb, Liu2021}. However, GAN-based approaches frequently encounter practical limitations, including mode collapse and unstable training, while reconstruction- and \textit{k}-space-based methods often require scanner-specific modifications. These limitations highlight the importance of image-domain methods that operate directly on magnitude images, which are universally available and readily integrated into existing clinical workflows without changes to acquisition hardware or vendor-specific reconstruction software.

In contrast, image-domain methods that operate directly on reconstructed magnitude images that are universally available in both retrospective studies and multi-center collaborations, enabling wide applicability without requiring changes to the acquisition hardware or reconstruction software. This characteristic makes image-based models particularly attractive as off-the-shelf solutions for clinical deployment. Furthermore, the use of image-domain methods allows integration with existing clinical pipelines for tasks such as segmentation, registration, and radiotherapy planning, where corrected magnitude images can be seamlessly substituted for corrupted ones. For these reasons, this study focused on the image domain, aiming to balance methodological innovation with clinical feasibility.

Recently, diffusion denoising probabilistic models (DDPMs) have revolutionized image generation techniques by markedly improving synthesis quality~\cite{10.5555/3495724.3496298} and have been adapted for various medical imaging tasks, including image synthesis~\cite{10167641, 10.1117/12.3047506}, denoising~\cite{WU2023104901}, MRI acceleration~\cite{https://doi.org/10.1002/mp.17675}, and the vision foundation model for MRI~\cite{wang2025triadvisionfoundationmodel, Sun2024, wang2025unifyingbiomedicalvisionlanguageexpertise}. DDPMs involve a forward process in which a Markov chain gradually transforms the input image into Gaussian noise, $\mathcal{N}(\bz, \bi)$, over a large number of steps, followed by a reverse process in which a neural network reconstructs the original image from the noisy data~\cite{luo2022understandingdiffusionmodelsunified}. Existing DDPM-based MoCo models concatenate the motion-corrupted image $ \y $ with $ \x_N \sim \mathcal{N}(\bz, \bi)$ and performing the backward diffusion to reconstruct the motion-free image $ \hat{\x} $ being similar to the ground truth image $ \x $~\cite{https://doi.org/10.1002/mp.16844, SARKAR2025108684, 10635444}. Although these methods achieved promising results, their reliance on numerous diffusion steps substantially increases the inference time. Additionally, initiating reconstruction from fully Gaussian noise $ \x_N \sim \mathcal{N}(\bz, \bi) $ might be suboptimal for MRI motion correction task (see Section~\ref{sec:ddpm_v01}).

In this study, we present \textbf{Res-MoCoDiff}, residual-guided efficient motion-correction denoising diffusion probabilistic model, a diffusion model that explicitly exploits the residual error between motion-free $ \x $ and motion-corrupted $ \y $ images (i.e., $ \rs = \y-\x $) in the forward diffusion process. Integrating this residual error into the diffusion process enables generation of noisy images at step $ N $ with a probability distribution closely matching that of the motion-corrupted images, specifically $ p(\x_N) \sim \mathcal{N}(\x; \y, \gamma^2 \bi) $. This approach offers two significant advantages: (1) enhanced reconstruction fidelity by avoiding the restrictive purely Gaussian prior assumption of conventional DDPMs, and (2) substantial computational efficiency, as the reverse diffusion process can be reduced to only four steps, substantially accelerating reconstruction times compared to traditional DDPMs. 



In summary, the main contributions of this study are as follows:

\begin{itemize}
	\item Res-MoCoDiff is an efficient diffusion model leveraging residual information, substantially reducing the diffusion process to just four steps.
	\item Res-MoCoDiff employs a novel noise scheduler that enables a more precise transition between diffusion steps by incorporating the residual error.
	\item Res-MoCoDiff replaces the attention layers with a Swin Transformer block.
	\item Extensive evaluation of Res-MoCoDiff is performed on both simulated (\textit{in-silico}) and clinical (\textit{in-vivo}) datasets covering various levels of motion-induced distortions.
\end{itemize}


\section{Materials and Methods}
\subsection{DDPM}\label{sec:ddpm_v01}
DDPMs are inspired by non-equilibrium thermodynamics and aim to approximate complex data distributions using a tractable distribution, such as a normal Gaussian distribution as a prior~\cite{10.5555/3045118.3045358}. Specifically, DDPMs employ a Markov chain consisting of two distinct processes: a forward diffusion and a backward (denoising) process. During the forward diffusion, the input image $ \x $ is gradually perturbed through a sequence of small Gaussian noise injections, eventually converging toward pure Gaussian noise $\mathcal{N}(\bz, \bi)$ after a large number of diffusion steps~\cite{luo2022understandingdiffusionmodelsunified,10.5555/3495724.3496298}. Conversely, the backward process employs a DL model to iteratively remove noise and reconstruct the original image from the Gaussian noise by approximating the reverse Markov chain of the forward diffusion.

In traditional DDPM implementations, this reconstruction (reverse diffusion) typically requires many iterative steps (often hundreds to thousands), significantly increasing the computational burden and limiting clinical applicability, especially in time-sensitive scenarios~\cite{luo2022understandingdiffusionmodelsunified, 10.5555/3495724.3496298}.


Formally, MoCo algorithms aim to recover an unknown motion-free image $\x \in \mathbb{R}^n$ from a motion-corrupted image $\y$ according to
\begin{equation}
	\y = \mathcal{A}(\x) + \mathbf{n} \in \mathbb{R}^n,
\end{equation}
where $\mathcal{A}$ denotes an unknown motion corruption operator and $\mathbf{n}$ represents additive noise. Since this inverse problem is ill-posed, it is essential to impose a regularization or prior assumptions to constrain the solution space. Without such constraints, multiple plausible solutions for $\x$ may be consistent with the observed data $\y$. From a Bayesian perspective, this regularization is introduced via a prior distribution $p(\x)$, which, when combined with the likelihood term $p(\y \vert \x)$, yields the posterior distribution:

\begin{equation}\label{eq:posterior_inverse_problem}
	\centering
	p(\x \vert \y) \propto p(\x)\,p(\y \vert \x)
\end{equation}


Traditional DDPMs typically assume a normal Gaussian prior, $p(\x) = \mathcal{N}(\bz, \bi)$. While mathematically convenient, this assumption might not be ideal for inverse problem tasks~\cite{10681246, NEURIPS2023_2ac2eac5} such as MRI motion correction tasks because it could encourage unrealistic reconstruction, introducing unwanted artifacts or image hallucinations, as suggested by recent studies~\cite{https://doi.org/10.1002/mp.16844}.

\subsection{Problem Formulation}
\begin{figure}[tbhp]
	\centering
	\includegraphics[width=\textwidth, draft=false]{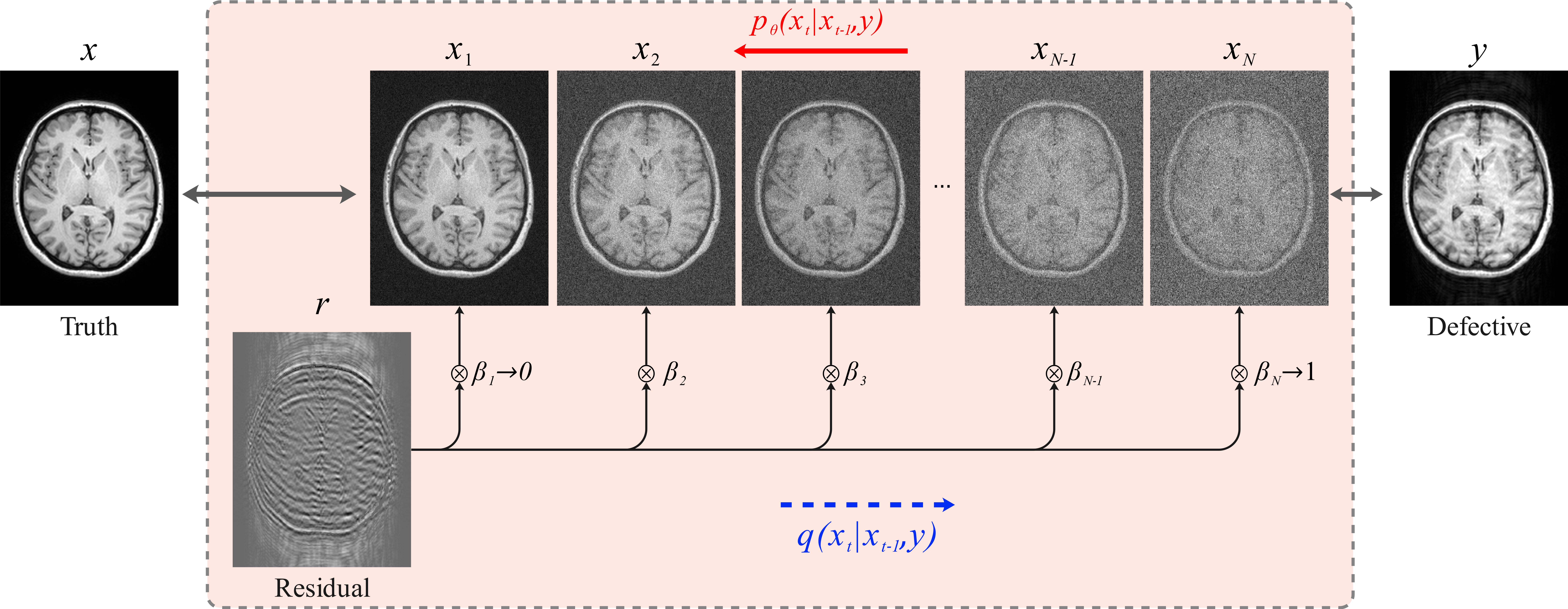}
	\caption{Flowchart of the Res-MoCoDiff approach. The forward process $ q(\x_{t} \vert \x_{t-1}, \y) $ employs a Markov chain to shift the residual error ($ \rs = \y - \x $), thus simulating the forward diffusion. The backward diffusion is also modeled via a Markov chain $ p_\theta(\x_t \vert \x_{t-1}, \y) $, where a DL model parametrized by $ \theta $ is trained to iteratively remove the noise and recover the original image.}
	\label{fig:flowchart_v01}
\end{figure}

Similar to conventional DDPMs, Res-MoCoDiff employs a Markov chain for both the forward and backward diffusion processes. However, it introduces a key modification: explicitly incorporating the residual error $ \rs $ between the motion-corrupted ($ \y $) and the motion-free ($ \x $) images into the forward diffusion process. This process is illustrated in \figurename~\ref{fig:flowchart_v01}.

\subsubsection{Forward Process} 
\paragraph{Res-MoCoDiff.}


Res-MoCoDiff employs a monotonically increasing shifting sequence $\{\beta_t\}_{t=1}^N$ to modulate the residual error $\rs$, starting with $\beta_1 \to 0$ and culminating in $\beta_N \to 1$, as illustrated in \figurename~\ref{fig:flowchart_v01}, where each forward step progressively integrates more of the residual into the motion-free image. The transition kernel for each forward step is given by

\begin{equation}\label{eq:shift_kernel_eq1}
	q(\x_t \vert \x_{t-1}, \y) = \mathcal{N}\left( \x_t; \x_{t-1} + \alpha_t \rs, \gamma^2\alpha_t  \bi \right) \quad \text{for} \quad t \in [0, N],
\end{equation}
where $\alpha_t = \beta_t - \beta_{t-1}$ and $\alpha_1 = \beta_1 \to 0$. The hyperparameter $\gamma$ enhances the flexibility of the forward process. Following a procedure similar to that described in \cite{luo2022understandingdiffusionmodelsunified,safari2025mrisuperresolutionreconstructionusing}, it can be shown that the marginal distribution of the data at a given time step $t$ from the input image $\x$ is

\begin{equation}\label{eq:marginal_distribution_input}
	q(\x_t \vert \x, \y) = \mathcal{N}\left( \x_t; \x + \beta_t \rs, \gamma^2\beta_t  \bi \right) \quad \text{for} \quad t \in [0, N],
\end{equation} 
where we denote the motion-free input image by $\x$, omitting the subscript (i.e., $\x_0$).

\begin{figure}[h]
	\centering
	\includegraphics[width=\textwidth, draft=false]{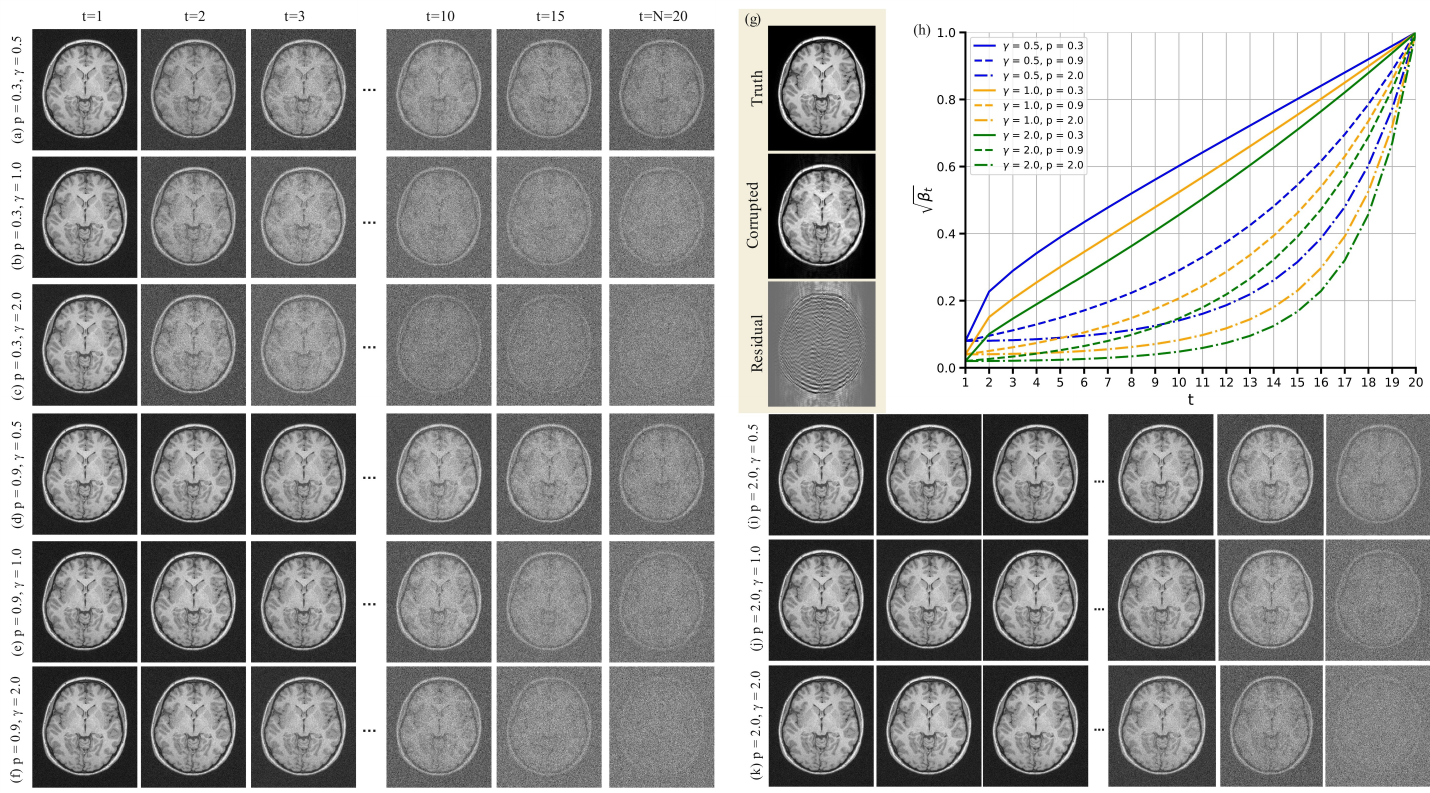}
	\caption{Illustration of the influence of hyperparameters on the forward diffusion process. Panels (a)–(c), (d)–(f), and (i)–(k) demonstrate how varying the hyperparameter $\gamma$ affects the noise level in the generated images $\x_t$ for different values of $p$, with higher $\gamma$ leading to stronger noise. Panels (a) and (d) specifically compare the effect of $p$ for a fixed $\gamma$. Panel (g) displays the ground truth motion-free image $\x$, the motion-corrupted image $\y$, and the residual error $\rs$. Panel (h) shows the evolution of $\sqrt{\beta_t}$ over the time steps $t$ for various hyperparameter combinations.}
	\label{fig:hyperparameters_v01}
\end{figure}

\paragraph{Noise scheduler.} We employ a non-uniform geometric noise scheduler, as proposed by Yue \textit{et al.}~\cite{NEURIPS2023_2ac2eac5}, to compute the shifting sequence $\{\beta_t\}_{t=1}^N$. Formally,

\begin{equation}\label{eq:noise_scheduler}
	\sqrt{\beta_t} = \beta_1 \exp\left(\frac{1}{2}\left[\frac{t - 1}{T - 1}\right]^p \log\frac{\beta_N}{\beta_1}\right) \quad \text{for} \quad t \in [2, N-1],
\end{equation}
where $p$ is a hyperparameter controlling the growth rate. As shown in \figurename~\ref{fig:hyperparameters_v01}, lower values of $p$ lead to greater noise levels in the images $\x_t$ across the forward diffusion steps. In addition, it is recommended to keep $\gamma \sqrt{\beta_1}$ sufficiently small to ensure $q(\x_1 \vert \x,\y) \simeq q(\x)$ (see \eqref{eq:marginal_distribution_input})~\cite{10.5555/3045118.3045358, 10.5555/3495724.3496298}. Hence, we set $\gamma \sqrt{\beta_1} = 0.04$ by choosing $\beta_1 = (0.04 / \gamma)^2$ and $\gamma = 2$. We also set $\beta_N = 0.999$ to satisfy the upper bound $\beta_N \to 1$. Unlike $p$, which modulates the rate at which noise accumulates, a larger $\gamma$ amplifies the overall noise level at each step. Panels (a)--(c), (d)--(f), and (i)--(k) of \figurename~\ref{fig:hyperparameters_v01} illustrate how different values of $p$ and $\gamma$ alter the forward diffusion process at various time steps $t$, while panel (g) depicts the ground truth $\x$, the motion-corrupted image $\y$, and the residual $\rs$. The corresponding noise scheduler curves for each hyperparameter combination are shown in panel (h).

\subsubsection{Backward process}

\begin{figure}
	\centering
	\includegraphics[width=\textwidth, draft=false]{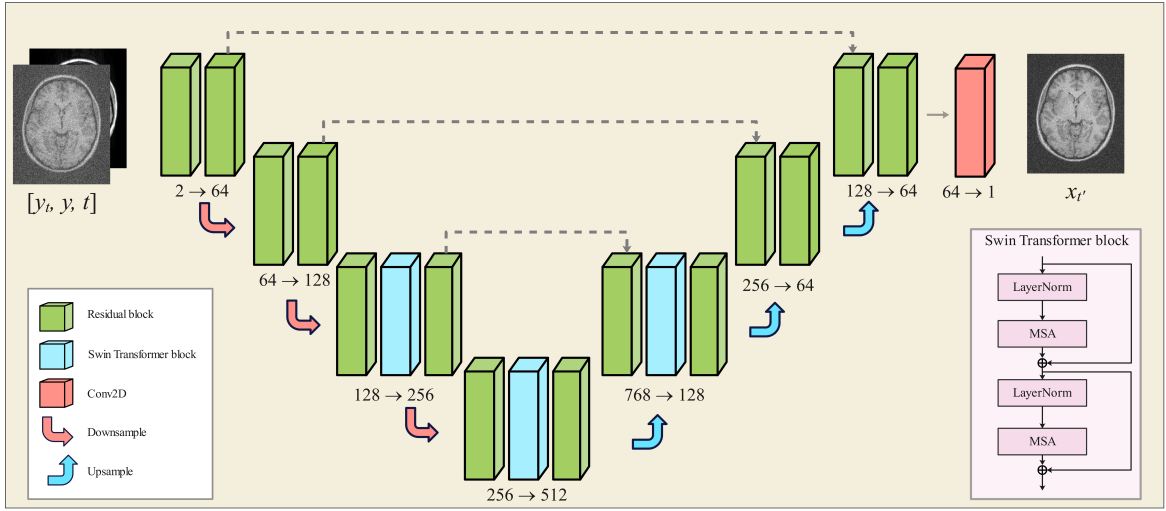}
	\caption{The Res-MoCoDiff network architecture. The inputs consist of a motion-corrupted image $\y$, a motion-free image $\x_t$ at a given time step $t$, and the corresponding time step information. The output is the estimated motion-free image $\x_{t^\prime}$ for $t^\prime < t$.}
	\label{fig:network}
\end{figure}

This process trains a DL model, parameterized by $\theta$, that employs a U-net backbone in which the conventional attention layers are replaced by Swin Transformer blocks~\cite{Liu_2021_ICCV} to improve generalization across different image resolutions~\cite{safari2025physicsinformeddeeplearningmodel}. The network architecture is depicted in \figurename~\ref{fig:network}.

The Res-MoCoDiff model is trained to estimate the posterior distribution $p_\theta(\x \vert \y)$ as follows:
\begin{equation}\label{eq:posterior_sampling_v01}
	\centering
	p_\theta (\x \vert \y) = \int p(\x_N \vert \y) \prod_{t=1}^{N} p_\theta (\x_{t-1} \vert \x_{t}, \y) d\x_{1:N}
\end{equation}
where $p(\x_N \vert \y) \approx \mathcal{N}(\y, \gamma^2 \bi)$, and $p_\theta (\x_{t-1} \vert \x_{t}, \y)$ denotes a DL model, parameterized by $\theta$, which approximates $\x_{t-1}$ given $\x_{t}$.

Following the conventional DDPM literature~\cite{NEURIPS2023_2ac2eac5, safari2025mrisuperresolutionreconstructionusing, luo2022understandingdiffusionmodelsunified, 10681246}, we assume that the reverse process follows a Gaussian distribution:
\begin{equation}\label{eq:gaussian_assumption}
	\centering
	p_\theta(\x_{t-1} \vert \x_{t}, y) = \mathcal{N}(\x_{t-1}; \mu_\theta(\x_{t}, \y, t), \bm{\Sigma}_\theta(\x_t, \y, t)), 
\end{equation}
where the parameters $\theta$ are optimized by minimizing the following evidence lower bound:

\begin{equation}\label{eq:lk_divergence}
	\centering
	\sum_{t=1}^{N} D_{KL}\left[q(\x_{t-1} \vert \x_{t}, x, y) \parallel p_\theta (\x_{t-1} \vert \x_t, y)\right],
\end{equation}
with $D_{KL}[\cdot \parallel \cdot]$ denoting the Kullback-Leibler divergence. Detailed derivations can be found in~\cite{NEURIPS2023_2ac2eac5, safari2025mrisuperresolutionreconstructionusing, 10681246}.

Based on \eqref{eq:shift_kernel_eq1} and \eqref{eq:marginal_distribution_input}, the target distribution $q(\x_{t-1} \vert \x_{t}, \x, \y)$ is given by:

\begin{equation}\label{eq:final_q_distributio}
	\centering
	q(\x_{t-1} \vert \x_{t}, x, y) =\mathcal{N}(\x_{t-1} ; \underset{\bm{\mu}_q}{\underbrace{\dfrac{\beta_{t-1}}{\beta_t} \x_{t} + \dfrac{\alpha_t}{\beta_t}\x}}, \underset{\bm{\Sigma}_q}{\underbrace{\gamma^2 \dfrac{\beta_{t-1}}{\beta_t} \alpha_t \bi}} ),
\end{equation}

Since $\bm{\Sigma}_q$ is independent of the inputs $\x$ and $\y$, we set $\bm{\Sigma}_\theta(\x_t,\y,t) = \bm{\Sigma}_q$, in accordance with previous works~\cite{10.5555/3495724.3496298, safari2025mrisuperresolutionreconstructionusing, 10681246}.

The mean parameter $\bm{\mu}_\theta(\x_t, \y, t)$ is modeled as follows:

\begin{equation}\label{eq:mean_parameter}
	\centering
	\bm{\mu}_\theta(\x_t, \y, t) = \dfrac{\beta_{t-1}}{\beta_t} \x_t + \dfrac{\alpha_t}{\beta_t} f_{\theta}(\x_t, \y, t)
\end{equation}
where $f_{\theta}(\cdot)$ denotes the DL model parameterized by $\theta$.

Under the assumption of a Gaussian kernel and a Markov chain, it can be shown that \eqref{eq:posterior_sampling_v01} can be optimized by minimizing the $\ell_{2}$ loss below,
\begin{equation}\label{eq:ell2_loss_alone}
	\centering
	\hat{\theta} = \underset{\theta}{\arg \min} \parallel f_{\theta}(\x_t, \y, t) - \x \parallel_2^2,
\end{equation}

Additionally, our experiments demonstrate that incorporating an $\ell_{1}$ regularizer can enhance high-resolution image reconstruction by promoting sparsity in the learned representations. The overall loss function is defined as:

\begin{equation}\label{eq:total_loss}
	\centering
	\mathcal{L}_\theta(\x_t,\y,t) = \parallel f_{\theta}(\x_t, \y, t) - \x \parallel_2^2 + \parallel f_{\theta}(\x_t, \y, t) - \x \parallel_1^1,
\end{equation}
with the effectiveness of the $\ell_{1}$ regularizer further validated in the ablation study in Section~\ref{sec:ablation_study}. The pseudo-codes for the training and sampling processes are provided in Algorithms~\ref{alg:training_process} and~\ref{alg:sampling_process}, respectively.

In our implementation, the training objective combined $\ell_1$ and $\ell_2$ losses with equal weighting. This design was chosen to balance the strengths of the two norms: the $\ell_2$ component penalizes large deviations and promotes overall pixel-level consistency, whereas the $\ell_1$ component preserves high-frequency features and sharper structural details. While $\ell_2$ loss alone can lead to overly smoothed reconstructions, the inclusion of $\ell_1$ mitigates this effect by encouraging sparsity and edge sharpness. This complementary behavior has been reported in prior medical image restoration studies and was confirmed in our ablation analysis (Section~\ref{sec:ablation_study}). No additional hyperparameter optimization of the relative weights was performed in this work.

In Res-MoCoDiff, the forward diffusion process was implemented with $N=20$ steps. However, the reverse process was reduced to only four steps due to the residual-guided formulation, which shifts the corrupted image distribution closer to the motion-free distribution. This substantially reduces the gap that the reverse diffusion must bridge. The number of reverse steps was selected empirically by evaluating different step counts on a validation set, where four steps provided the optimal trade-off between reconstruction quality and efficiency.

\begin{algorithm}[t]
\caption{Training process}\label{alg:training_process}
\begin{algorithmic}
	\Require motion-free dataset $\mathcal{T}$, motion-corrupted dataset $\mathcal{T}_c$
	\Repeat 
	
		$\x \sim \mathcal{T}$, $\y \sim \mathcal{T}_c$
	
		$t \sim $ Uniform$(\{1, ..., N\})$
		
		$\x_t \sim q(\x_t \vert \x, \y,t)$\Comment{Given in \eqref{eq:marginal_distribution_input}}
		
		Take a gradient descent step on $\nabla\mathcal{L}_\theta(\x_t,\y,t)$\Comment{Given in \eqref{eq:total_loss}}
	\Until converged
\end{algorithmic}
\end{algorithm}

\begin{algorithm}[h]
	\caption{Sampling process}\label{alg:sampling_process}
	\begin{algorithmic}
		\Require motion-corrupted image $\y$; number of steps $N=4$; noise scheduler $\{\beta_t\}_{t=1}^N$ with $\beta_N = 0.999$, 
		$\beta_1 = (0.04/\gamma)^2$, growth rate $p=0.3$, and 
		$\alpha_t = \beta_t - \beta_{t-1}$; noise scaling $\gamma=2$
		
		$\x_N \sim \mathcal{N}(\x_N; \y, \gamma^2 \beta_N \bi)$
		
		\For{t = N, ..., 1}
		
		$\bm{\epsilon} \sim \mathcal{N}(\bm{\epsilon}; \bz, \bi)$ if $t > 1$ else $\bm{\epsilon} = 0$
		
		$ \bm{\mu}_\theta =  \dfrac{\beta_{t-1}}{\beta_t} \x_t + \dfrac{\alpha_t}{\beta_t} f_{\theta}(\x_t, \y, t) $ \Comment{Given in \eqref{eq:mean_parameter}}
		
		$\x_{t-1} = \bm{\mu}_\theta + \gamma \sqrt{\dfrac{\beta_{t-1} \alpha_t}{\beta_t}} \bm{\epsilon} $
		\EndFor
		
	\end{algorithmic}
\end{algorithm}


Res-MoCoDiff was implemented in \texttt{PyTorch} (version 2.5.1) and executed on an NVIDIA A100 with 80 GB GPU RAM. The model was trained for 100 epochs with a batch size of 32. Optimization was performed using the RAdam optimizer~\cite{DBLP:conf/iclr/LiuJHCLG020} in conjunction with a cosine annealing learning rate scheduler~\cite{DBLP:conf/iclr/LoshchilovH17}, with an initial learning rate of $2 \times 10^{-4}$ and a minimum learning rate of $2 \times 10^{-5}$. A warm-up phase comprising 5,000 steps was employed prior to transitioning to the cosine schedule to stabilize early training dynamics.

\subsection{Patient Data Acquisition and Data Pre-processing}
This study utilizes two publicly available datasets, namely the IXI dataset (\url{https://brain-development.org/ixi-dataset/}) and the movement-related artifacts (MR-ART) dataset from Open-Neuro~\cite{Narai2022}, to train and evaluate our models.

The IXI dataset comprises 580 cases of T1-weighted (T1-w) brain MRI images. We partitioned the dataset into two non-overlapping subsets: a training set consisting of 480 cases (54,160 slices) and a testing set comprising 100 cases (11,980 slices). We adapted the motion simulation technique of Duffy \textit{et al.}~\cite{DUFFY2021117756} to generate an \textit{in-silico} dataset with varying levels of motion artifacts including high, moderate, and minor by perturbing 15, 10, and 7 \textit{k}-space lines along a phase encoding direction, respectively. Random slabs, with widths ranging between three and seven \textit{k}-space lines, were selected along the phase encoding direction and were subjected to rotational perturbations of $\pm 7^\circ$ and translational shifts of $\pm 5$ mm.

Additionally, model performance on \textit{in-vivo} data was evaluated using the MR-ART T1-w brain MRI dataset, which comprises 148 cases (95 females and 53 males). This dataset includes three types of images: ground truth motion-free images, motion-corrupted images with a low level of distortion (level 1), and motion-corrupted images with a high level of distortion (level 2). Rigid brain image registration was performed using FSL-FLIRT~\cite{JENKINSON2002825, JENKINSON2001143} to compensate for misalignment between the motion-free and motion-corrupted images.

It is important to note that our study was conducted in the image domain, rather than in \textit{k}-space. This choice was made because most publicly available datasets and many clinical archives provide only reconstructed magnitude images, not raw \textit{k}-space data. Working in the image domain therefore enables broader applicability of Res-MoCoDiff to retrospective studies and multi-institutional data, where access to raw acquisition information is not readily available.

\subsection{Quantitative and Statistical Analysis}
We compared our model against benchmark approaches, including CycleGAN~\cite{CycleGAN2017}, Pix2pix~\cite{isola2017image}, and a conventional DDPM variant that employs a vision transformer backbone~\cite{Pan_2023}.

To quantitatively assess the performance of the models in removing brain motion artifacts, we reported three metrics: normalized mean squared error (NMSE), structural similarity index measure (SSIM)~\cite{1284395}, and peak signal-to-noise ratio (PSNR). Lower NMSE values indicate better performance, although NMSE may favor solutions with increased blurriness~\cite{Zhang_2018_CVPR}. SSIM ranges from -1 to 1, with a value of 1 representing optimal structural similarity between the reconstructed and ground truth images. Likewise, a higher PSNR denotes improved performance and is more aligned with human perception due to its logarithmic scaling~\cite{https://doi.org/10.1002/mp.17079}. The quantitative metrics were computed using the \texttt{PIQ} library (\url{https://piq.readthedocs.io/en/latest}, version 0.8.0)~\cite{kastryulin2022piq} with its default parameters.

\section{Results}

This section presents both qualitative and quantitative results for the \textit{in-silico} and \textit{in-vivo} datasets. In addition, an ablation study is conducted to quantify the contribution of each component of the proposed Res-MoCoDiff model.

\subsection{Qualitative results}

\begin{figure}
	\centering
	\includegraphics[width=0.9\textwidth, draft=false]{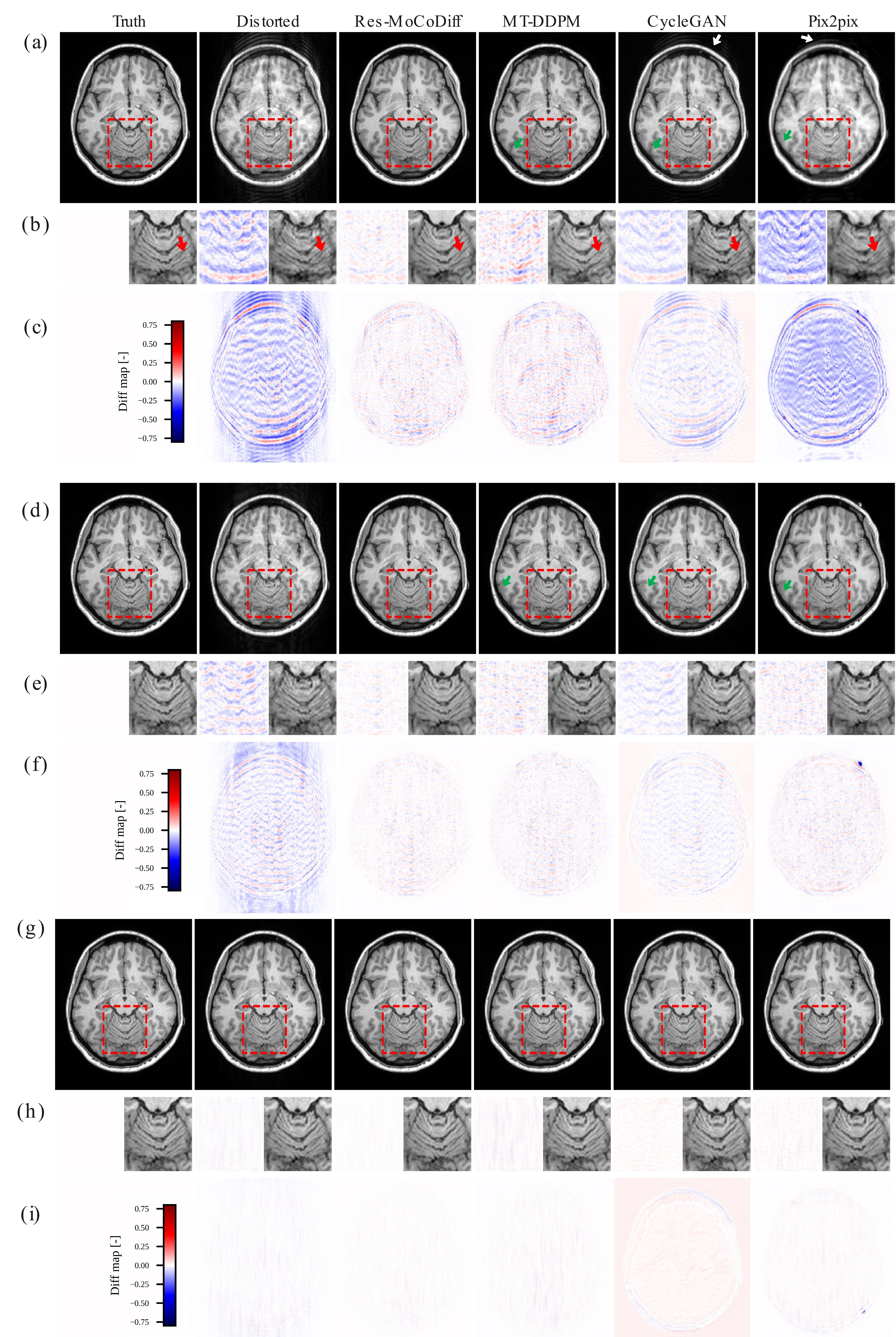}
	\caption{Qualitative results for the \textit{in-silico} dataset are shown. Panels (a)-(c), (d)-(f), and (g)-(i) illustrate the outcomes for heavy, moderate, and minor distortion levels, respectively. The white and green arrows in panels (a) and (d) indicate ringing artifacts inside the skull and ghosting of bright fat tissue outside the skull, respectively, while panels (b), (e), and (h) present zoomed-in views of the regions highlighted by the red boxes.}
	
	\label{fig:qualitative_inSilico}
\end{figure}

The motion artifacts observed in the motion-corrupted images confirm that our simulation procedure successfully reproduces both ringing artifacts inside the skull and ghosting of bright fat tissue outside the skull, as indicated by the white and green arrows in Figures~\ref{fig:qualitative_inSilico}(a) and (d). Notably, the zoomed-in regions in \figurename~\ref{fig:qualitative_inSilico}(b) illustrate that Res-MoCoDiff preserves fine structural details more effectively than the comparative methods. Furthermore, the pixel-level distortion maps in Figures~\ref{fig:qualitative_inSilico}(c), (f), and (i) underscore the superior artifact removal achieved by Res-MoCoDiff.

\begin{figure}[h]
	\centering
	\includegraphics[width=0.9\textwidth, draft=false]{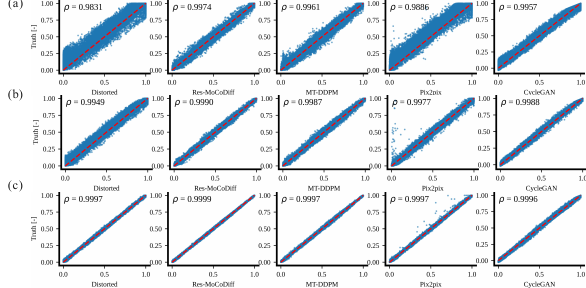}
	\caption{Pixel-wise correlations for the \textit{in-silico} dataset are shown. Panels (a)-(c) display the corresponding pixel-wise correlation plots for heavy, moderate, and minor distortion levels.}
	
	\label{fig:qualitative_inSilico_correlations}
\end{figure}

Although our approach demonstrates a generally robust ability to preserve detailed structures, a few residual ringing artifacts remain (highlighted by arrows in \figurename~\ref{fig:qualitative_inSilico}(d)) for the moderate distortion level. For the minor distortion level, the overall performance among all methods is similar in mitigating motion artifacts, as shown in \figurename~\ref{fig:qualitative_inSilico}(g)-(i). Finally, the pixel-wise correlation plots in \figurename~\ref{fig:qualitative_inSilico_correlations}(a)-(c) confirm the qualitative findings: Res-MoCoDiff attained Pearson correlation coefficients of $\rho = 0.9974$, $\rho = 0.9990$, and $\rho = 0.9999$ for high, moderate, and minor distortion levels, respectively, surpassing the second-best MT-DDPM method, which yields $\rho = 0.9961$, $\rho = 0.9987$, and $\rho = 0.9997$.

\begin{figure}[tbph]
	\centering
	\includegraphics[width=\textwidth, draft=false]{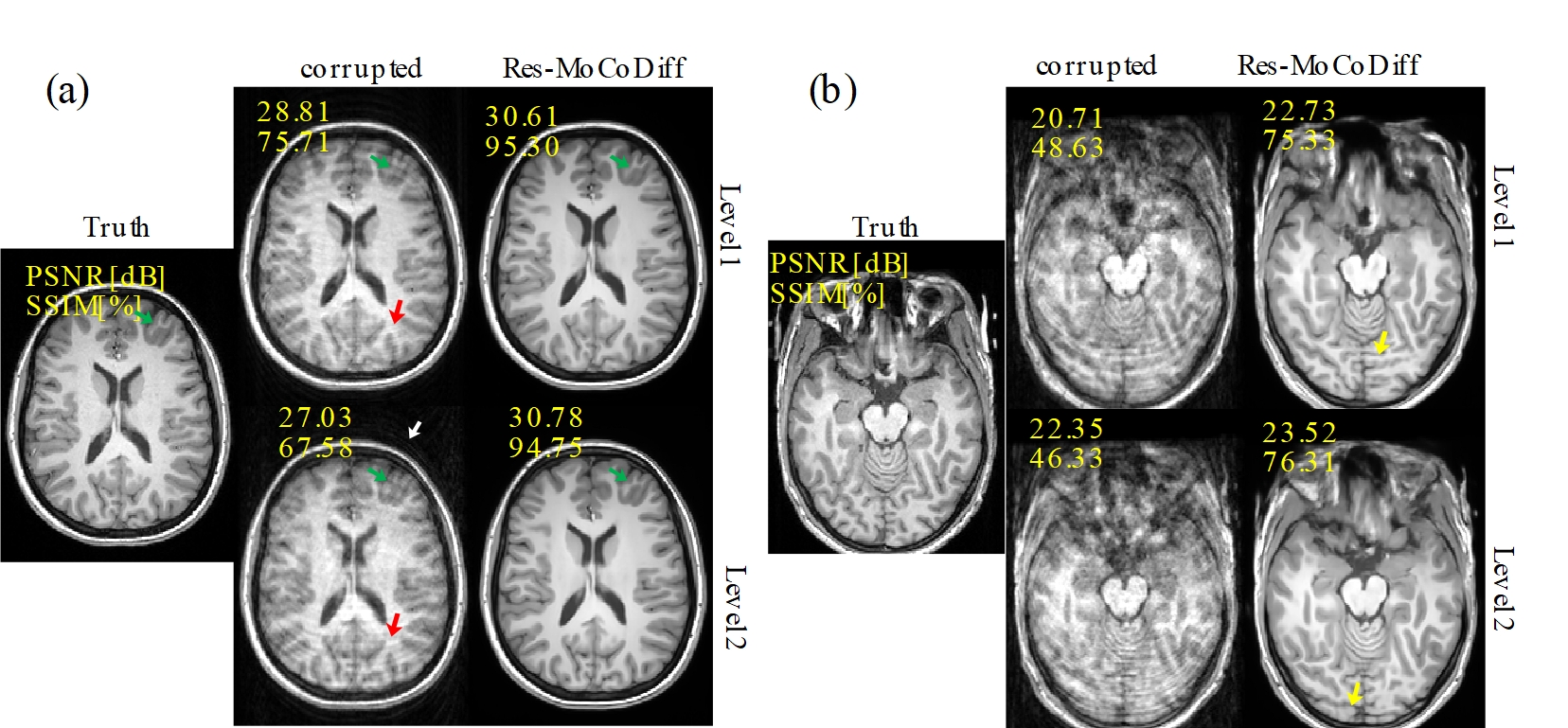}
	\caption{Qualitative results for the \textit{in-vivo} MR-ART dataset, illustrating both optimal (a) and suboptimal (b) reconstructions. Green and red arrows indicate ringing artifacts inside the skull, while the white arrow highlights ghosting of bright fat tissue outside the skull. Yellow arrows denote regions where Res-MoCoDiff hallucinated structures due to insufficient soft-tissue contrast in the corrupted inputs.}
	\label{fig:qualitative_invivo}
\end{figure}

Qualitative results for the \textit{in-vivo} MR-ART dataset are presented in Figure~\ref{fig:qualitative_invivo}, where both optimal and suboptimal reconstructions are illustrated. In panel (a), corresponding to cases with recoverable motion corruption, red arrows indicate ringing artifacts inside the skull, and a white arrow highlights ghosting of bright fat tissue outside the skull. The green arrows denote regions where Res-MoCoDiff successfully restores fine structural details, achieving improvements from 28.81 dB and 75.71 (motion-corrupted) to 30.61 dB and 95.30 (Res-MoCoDiff) in PSNR and SSIM for Level 1, and from 27.03 dB and 67.58 to 30.78 dB and 94.75 for Level 2. In panel (b), however, we show suboptimal reconstructions in which Res-MoCoDiff failed to fully recover the anatomical structures. As indicated by the yellow arrows, the model produced hallucinated features when the motion-corrupted input lacked sufficient soft-tissue contrast, underscoring the inherent difficulty of reconstructing details that are not present in the corrupted images.

\subsection{Quantitative results}

\begin{figure}[h]
	\centering
	\includegraphics[width=\textwidth, draft=false]{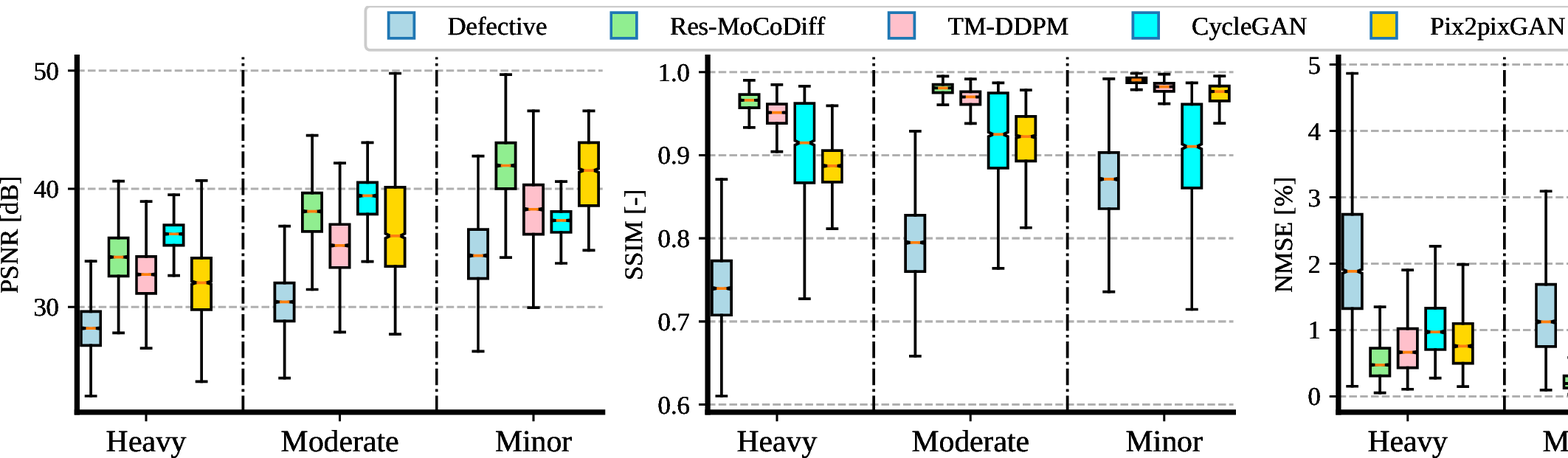}
	\caption{Boxplots of PSNR, SSIM, and NMSE metrics across different motion artifact levels for the \textit{in-silico} IXI dataset.}
	\label{fig:boxplot}
\end{figure}

As illustrated in \figurename~\ref{fig:boxplot}, motion corruption progressively reduces PSNR and SSIM across minor, moderate, and heavy distortion levels, reaching average values of $(34.62 \pm 3.25\text{ dB},\,0.87 \pm 0.05)$, $(30.46 \pm 2.48\text{ dB},\,0.79 \pm 0.05)$, and $(28.14 \pm 2.20\text{ dB},\,0.74 \pm 0.05)$, respectively. This negative trend confirms that increased distortion degrades image quality. Conversely, NMSE values rise from $0.56 \pm 0.43 \%$ to $1.33 \pm 0.84 \%$ and $2.94 \pm 1.79 \%$ as the distortion intensifies. These results are further detailed in \tablename~\ref{tab:quantitative_values}, which shows that our proposed Res-MoCoDiff method consistently achieves higher PSNR and SSIM, as well as lower NMSE, compared with the benchmark approaches at all distortion levels.

\begin{table}[b]
	\centering
	\caption{Quantitative Metrics demonstrated in mean$_{\pm \text{std}}$ across different motion artifact levels of the \textit{in-silico} dataset are summarized. The arrows indicate the direction of better performance. Bold indicates the best values.}
	\label{tab:quantitative_values}
	\resizebox{\textwidth}{!}{%
		\begin{tabular}{lllllll}
			\hline
			Metrics& Distortion level           
			                  & Corrupted                     & Pix2Pix           & CycleGAN            & MT-DDPM                & Res-MoCoDiff (ours)    \\ \hline
			\multirow{3}{*}{PSNR [dB] $\uparrow$}
			& Minor            & $34.62_{\pm 3.25}$           & $37.03_{\pm 1.65}$     & $41.21_{\pm 2.91}$          & $38.25_{\pm 3.08}$                  & $\bm{41.91_{\pm 2.94}}$ \\
			& Moderate         & $30.46_{\pm 2.48}$           & $37.16_{\pm 4.86}$     & $\bm{38.96_{\pm 2.33}}$      & $35.10_{\pm 2.64}$                  & $\underline{37.97_{\pm 2.39}}$ \\
			& Heavy            & $28.14_{\pm 2.20}$           & $31.95_{\pm 3.01}$     & $\bm{35.93_{\pm 1.48}}$       & $32.65_{\pm 2.30}$                  & $\underline{34.15_{\pm 2.42}}$ \\ \cdashline{1-7}
			\multirow{3}{*}{SSIM [-] $\uparrow$}  
			& Minor            & $0.87_{\pm 0.05}$            & $0.91_{\pm 0.07}$      & $0.97_{\pm 0.01}$              & $\underline{0.98_{\pm 0.01}}$      & $\bm{0.99_{\pm 0.00}}$    \\
			& Moderate         & $0.79_{\pm 0.05}$            & $0.92_{\pm 0.05}$      & $0.92_{\pm 0.05}$              & $\underline{0.98_{\pm 0.01}}$    & $\bm{0.98_{\pm 0.01}}$    \\
			& Heavy            & $0.74_{\pm 0.05}$            & $0.89_{\pm 0.03}$      & $0.91_{\pm 0.06}$              & $\underline{0.95_{\pm 0.02}}$   & $\bm{0.96_{\pm 0.01}}$    \\ \cdashline{1-7}
			\multirow{3}{*}{NMSE [\%] $\downarrow$}  
			& Minor            & $0.56_{\pm 0.43}$            &     $0.18_{\pm 0.13}$  & $\underline{0.15_{\pm 0.18}}$   & $0.24_{\pm 0.20}$                  & $\bm{0.10_{\pm 0.09}}$    \\
			& Moderate         & $1.33_{\pm 0.84}$            & $0.57_{\pm 0.43}$      & $0.60_{\pm 0.60}$              & $\underline{0.47_{\pm 0.32}}$      & $\bm{0.24_{\pm 0.16}}$    \\
			& Heavy            & $2.94_{\pm 1.79}$            &  $1.14_{\pm 0.69}$     & $0.88_{\pm 0.54}$              & $\underline{0.81_{\pm 0.56}}$      & $\bm{0.58_{\pm 0.40}}$    \\ \hline
		\end{tabular}%
	}
\end{table}

As shown in \figurename~\ref{fig:boxplot} and \tablename~\ref{tab:quantitative_values}, Res-MoCoDiff consistently achieves the lowest NMSE across all distortion levels. For minor distortion, Res-MoCoDiff outperforms all comparative methods in PSNR ($41.91 \pm 2.94\,\text{dB}$) and SSIM ($0.99 \pm 0.00$), while also obtaining the lowest NMSE ($0.10 \pm 0.09\%$). CycleGAN provides the second-best NMSE ($0.15 \pm 0.18\%$) at this level, although it achieves a lower PSNR ($41.21 \pm 2.91\,\text{dB}$) and SSIM ($0.97 \pm 0.01$) compared with Res-MoCoDiff.

For moderate distortion, Res-MoCoDiff achieves the highest SSIM ($0.98 \pm 0.01$) and the lowest NMSE ($0.24 \pm 0.16\%$). CycleGAN attains the best PSNR ($38.96 \pm 2.33\,\text{dB}$) for this distortion level, but its SSIM ($0.92 \pm 0.05$) and NMSE ($0.60 \pm 0.60\%$) remain below Res-MoCoDiff’s performance. MT-DDPM ranks as the second-best method in NMSE ($0.47 \pm 0.32\%$), highlighting its competitive artifact-reduction capabilities, although it still trails Res-MoCoDiff.

At the heavy distortion level, Res-MoCoDiff again secures the best SSIM ($0.96 \pm 0.01$) and NMSE ($0.58 \pm 0.40\%$), whereas CycleGAN achieves the highest PSNR ($35.93 \pm 1.48\,\text{dB}$). Nonetheless, CycleGAN’s SSIM ($0.91 \pm 0.06$) and NMSE ($0.88 \pm 0.54\%$) are notably worse than those of Res-MoCoDiff, indicating that a higher PSNR alone may not guarantee superior structural fidelity or overall artifact removal. Across all three distortion levels, the boxplots reveal that Res-MoCoDiff’s performance distribution is consistently shifted toward higher PSNR and SSIM and lower NMSE values compared with the other methods, underscoring its robustness in mitigating motion artifacts.

In the \textit{in-vivo} MR-ART dataset, the original motion-corrupted images at distortion levels 1 and 2 yield NMSE, SSIM, and PSNR values of $(2.56 \pm 2.38\%,$ $0.74{\pm0.10},$ $28.61{\pm2.85}\, \text{dB})$ and $(3.30{\pm2.76}\%,\, 0.72 \pm 0.10,\, 27.51 \pm 2.83)$, respectively. Res-MoCoDiff reduces NMSE by $33.21\%$ (to $1.71{\pm1.49}$) for level~1 and $37.41\%$ (to $2.07{\pm1.79}$) for level~2. Additionally, it raises SSIM by $23.60\%$ (to $0.92 \pm 0.05$) for level~1 and $25.58\%$ (to $0.91 \pm 0.05$) for level~2, while also improving PSNR by $6.26\%$ (to $30.40 \pm 2.90\,\text{dB}$) for level~1 and $7.71\%$ (to $29.63 \pm 2.97\,\text{dB}$) for level~2. These gains underline the robust performance of Res-MoCoDiff for both \textit{in-silico} and \textit{in-vivo} motion artifact correction.

%

\subsection{Ablation Study}\label{sec:ablation_study}
We conducted an ablation study to quantify the contributions of the $\ell_2$ loss defined in \eqref{eq:ell2_loss_alone} and the combined $\ell_1 + \ell_2$ loss specified in \eqref{eq:total_loss} to the overall performance of the proposed method across minor, moderate, and heavy distortion levels. \tablename~\ref{tab:ablation} summarizes the PSNR, SSIM, and NMSE metrics obtained under the two training scenarios.

When training with only the $\ell_2$ loss, the model achieved a PSNR of $40.44 \pm 2.77$ dB, an SSIM of $0.99 \pm 0.01$, and an NMSE of $0.14 \pm 0.11\%$ for minor distortion. For moderate distortion, the performance was $37.21 \pm 2.26$ dB in PSNR, $0.97 \pm 0.01$ in SSIM, and $0.28 \pm 0.18\%$ in NMSE, while for heavy distortion the corresponding values were $33.74 \pm 2.15$ dB, $0.95 \pm 0.01$, and $0.61 \pm 0.37\%$, respectively.

In contrast, when the model was trained using the complete Res-MoCoDiff strategy that incorporates both the $\ell_1$ and $\ell_2$ losses, performance improvements were observed consistently across all distortion levels. For minor distortion, the PSNR increased to $41.91 \pm 2.94$ dB (an improvement of $3.63\%$), SSIM improved marginally to $0.99 \pm 0.00$ (an increase of $0.42\%$), and the NMSE was reduced to $0.10 \pm 0.09\%$, corresponding to a reduction of $26.29\%$. For moderate distortion, the PSNR improved to $37.97 \pm 2.39$ dB (a $2.04\%$ increase), the SSIM increased to $0.98 \pm 0.01$ (an improvement of $0.48\%$), and the NMSE decreased to $0.24 \pm 0.16\%$, a reduction of $14.74\%$. For heavy distortion, the use of the combined loss resulted in a PSNR of $34.15 \pm 2.42$ dB (an improvement of $1.21\%$), an SSIM of $0.96 \pm 0.01$ (an increase of $0.95\%$), and an NMSE of $0.58 \pm 0.40\%$, corresponding to a reduction of $5.73\%$.

These findings indicate that the inclusion of the $\ell_1$ regularizer is instrumental in reducing pixel-level errors and preserving fine structural details, thereby contributing significantly to the overall performance of the method. The improvements, particularly in NMSE, underscore the efficacy of the combined loss function in mitigating residual errors and enhancing the robustness of motion artifact correction across varying levels of distortion.

\begin{table}[tbhp]
	\centering
	\caption{The ablation study results are summarized. The arrows indicate the direction of better performance. The numbers inside the parentheses in \textcolor{red}{red} are the improvement of the complete Res-MoCoDiff compared with the other training scenarios that only used $ \ell_{2} $ loss in training.}
	\label{tab:ablation}
	\begin{tabular}{lllll}
		\hline
		Scenarios& Distortion level & PSNR [dB] $ \uparrow $ & SSIM [-] $ \uparrow $ & NMSE [\%] $ \downarrow $ \\ \hline
		& Minor            & $ 40.44_{\pm 2.77} $ & $ 0.99_{\pm 0.01} $ & $ 0.14_{\pm 0.11} $ \\
		& Moderate         & $ 37.21_{\pm 2.26} $ & $ 0.97_{\pm 0.01} $ & $ 0.28_{\pm 0.18} $ \\
		\multirow{-3}{*}{$ \ell_2 $} & Heavy & $ 33.74_{\pm 2.15} $ & $ 0.95_{\pm 0.01} $ & $ 0.61_{\pm 0.37} $ \\  \cdashline{1-5}
		& Minor & $ 41.91_{\pm 2.94 \textcolor{red}{(3.63\%)}} $ & $ 0.99_{\pm 0.00\textcolor{red}{(0.42\%)}} $ & $ 0.10_{\pm 0.09\textcolor{red}{(-26.29\%)}} $ \\
		& Moderate & $ 37.97_{\pm 2.39\textcolor{red}{(2.04\%)}} $ & $ 0.98_{\pm 0.01\textcolor{red}{(0.48\%)}} $ & $ 0.24_{\pm 0.16\textcolor{red}{(-14.74\%)}} $ \\
		\multirow{-3}{*}{\begin{tabular}[c]{@{}l@{}}$ \ell_2 + \ell_1 $ (Res-\\ MoCoDiff)\end{tabular}} & Heavy & $ 34.15_{\pm 2.42\textcolor{red}{(1.21\%)}} $ & $ 0.96_{\pm 0.01\textcolor{red}{(0.95\%)}} $ & $ 0.58_{\pm 0.40\textcolor{red}{(-5.73\%)}} $ \\ \hline
	\end{tabular}
\end{table}

\section{Discussion}
MRI is a versatile imaging modality that provides excellent soft-tissue contrast and valuable physiological information. However, the prolonged acquisition times inherent to MRI increase the likelihood of patient motion, which in turn manifests as ghosting and ringing artifacts. Although the simplest solutions to mitigate motion artifacts involve repeating the scan or employing motion tracking systems, these approaches impose additional costs and burdens on the clinical workflow.

In this study, we proposed Res-MoCoDiff, an efficient denoising diffusion probabilistic model designed to reconstruct motion-free images. By leveraging a residual error shifting mechanism (illustrated in \figurename~\ref{fig:flowchart_v01}), our method performs the sampling process in only four steps (see Algorithm~\ref{alg:sampling_process}), thereby facilitating its integration into current clinical practices. Unlike conventional DDPMs that require hundreds of reverse steps, Res-MoCoDiff leverages residual error shifting to bring the corrupted distribution closer to the motion-free image distribution at the terminal diffusion step. This enables accurate image restoration in only four reverse steps. Our validation experiments showed that using more than four steps did not yield significant improvements, confirming the efficiency of this reduced-step sampling. Notably, Res-MoCoDiff achieves an average sampling time of 0.37 seconds per batch of two image slices, which is substantially lower than the 101.74 seconds per batch required by the conventional TM-DDPM approach.

Our motion simulation technique effectively generates realistic artifacts, including ringing within the skull and ghosting of bright fat tissue outside the skull, as indicated by the white arrows in \figurename~\ref{fig:qualitative_inSilico} for the \textit{in-silico} dataset and similarly in \figurename~\ref{fig:qualitative_invivo} for the \textit{in-vivo} dataset. This capability underscores the potential of our simulation framework to closely mimic the clinical appearance of motion artifacts.

Extensive qualitative and quantitative evaluations on both \textit{in-silico} and \textit{in-vivo} datasets demonstrate the superior performance of Res-MoCoDiff in removing motion artifacts across different distortion levels. While comparative models often leave residual artifacts, particularly under heavy and moderate motion conditions, Res-MoCoDiff consistently eliminates these imperfections, as highlighted by the green arrows in \figurename~\ref{fig:qualitative_inSilico}. Moreover, the proposed method excels at recovering fine structural details, resulting in higher pixel-wise correlations (see \figurename~\ref{fig:qualitative_inSilico_correlations}(a)-(c)). Although our method achieves the second highest PSNR among the evaluated techniques (see \tablename~\ref{tab:quantitative_values}), its overall improvements in SSIM and NMSE, together with the perceptually superior image quality, underscore its clinical efficacy.


The inclusion of an $\ell_1$ regularizer during training further enhanced image sharpness by reducing NMSE, which is particularly important given that higher NMSE is often associated with blurry reconstructions. This improvement was consistent with the increases observed in both PSNR and SSIM, resulting in images that were structurally closer to the ground truth~\cite{Safari2023}.

While $\ell_2$ loss is widely used in medical image restoration to reduce global pixel-level error, it can lead to oversmoothed reconstructions. By contrast, the $\ell_1$ term preserves high-frequency information and improves edge sharpness, thereby producing more realistic structures. The complementary behavior of $\ell_1$ and $\ell_2$ has been reported in both general image restoration~\citep{zhang2018unreasonable} and MRI artifact removal~\citep{liu2021learning}. Our ablation study (Table~\ref{tab:ablation}) confirmed these benefits, showing that the combined loss reduced NMSE while enhancing perceptual quality compared with $\ell_2$ alone.

A further consideration in designing Res-MoCoDiff is our decision to operate in the image domain rather than in \textit{k}-space. While \textit{k}-space methods can, in principle, provide direct access to raw acquisition information, their use in clinical practice is limited by several practical factors. Raw \textit{k}-space data are not routinely stored in most hospital archives or large-scale public repositories, and reconstruction pipelines differ substantially across scanner vendors and software versions, which complicates reproducibility. In addition, implementing \textit{k}-space motion correction often requires vendor-specific software environments that are not widely accessible. By contrast, reconstructed magnitude images are universally available and can be directly integrated into existing clinical workflows. Our image-domain design therefore prioritizes generalizability and clinical feasibility, allowing Res-MoCoDiff to serve as an off-the-shelf solution that can be applied retrospectively across a wide range of studies and institutions.

We focused on T1-w brain MRI because this sequence is both highly susceptible to motion artifacts and is central to Radiation Oncology workflows~\cite{DePietro2024}. This emphasis ensures that our evaluation addresses a clinically important sequence with high impact on both diagnostic and therapeutic decision making.

This study has several limitations that warrant discussion. First, in severely degraded \textit{in-vivo} cases where the motion-corrupted input lacks soft-tissue contrast, Res-MoCoDiff may hallucinate anatomical details that are not consistent with the ground truth. This limitation reflects the fact that when essential structural information is absent from the input, the model cannot reliably recover it. Future work should therefore explore strategies such as incorporating multi-contrast MRI, embedding stronger physics-informed priors, or using uncertainty quantification to identify high-risk reconstructions. Second, our work was restricted to brain MRI, and other sequences such as diffusion-weighted echo-planar imaging (EPI) and high-resolution T2w imaging were not explored. Motion in brain MRI is typically irregular, unpredictable, and of relatively small amplitude, and the residual-guided design of Res-MoCoDiff is well suited for these conditions. However, larger and non-rigid motion events, such as sneezing or swallowing during long 3D acquisitions, may introduce more complex artifacts that are not fully addressed by the current framework. In addition, our motion simulation followed the established approach of Duffy et al.~\cite{DUFFY2021117756}, which reproduces realistic ghosting and ringing but does not explicitly account for the increased probability of motion during longer scans. Furthermore, it should also be clarified that this study addresses retrospective motion artifact correction in structural brain MRI, which is distinct from physiological motion correction approaches (e.g., respiratory or cardiac compensation) used in free-breathing acquisitions. To extend Res-MoCoDiff to these broader scenarios, retraining on sequence-specific data may be required, and future studies could explore whether architectural modifications would further improve performance in modeling non-rigid motion. Addressing these extensions will be important for establishing the generalizability of the framework across diverse imaging protocols and clinical applications. Finally, future work should also explore integration with motion modeling or hybrid prospective-retrospective strategies to improve robustness in more complex scenarios.

\section{Conclusion}

Res-MoCoDiff represents a significant advancement in motion artifact correction for MRI. Its rapid processing speed and robust performance across a range of distortion levels make it a promising candidate for clinical adoption, potentially reducing the need for repeated scans and thereby improving patient throughput and diagnostic and treatment efficiency. Future work will focus on further optimizing the model, exploring its application to other imaging modalities, and validating its performance in larger, multi-center clinical studies.

\section*{Conflicts of interest}
There are no conflicts of interest declared by the authors.

\section*{Acknowledgment}
This research is supported in part by the National Institutes of Health under Award Numbers R56EB033332, R01DE033512, and R01CA272991.

\section*{Data availability}

The datasets used in this study are publicly available. The IXI dataset can be accessed at \url{https://brain-development.org/ixi-dataset/}, and the MR-ART dataset is available through OpenNeuro \url{https://openneuro.org/datasets/ds004173/versions/1.0.2}.

\bibliographystyle{unsrt}
\bibliography{./Res_MoCoDiff.bib}      

\end{document}